\documentclass[12pt,conference,letter]{IEEEtran}

\usepackage{flushend}
\usepackage{float}
\usepackage{cite}
\usepackage{graphicx}

\begin{document}

\title{ Face Recognition as a Method of Authentication in a Web-Based System }
\pdfinclusioncopyfonts=1
	\author{\IEEEauthorblockN{Ben Wycliff Mugalu}
\IEEEauthorblockA{\textit{Dept. Electrical and Computer Engineering} \\
\textit{Makerere University}\\
ben12wycliff@gmail.com}
\and
\IEEEauthorblockN{Rodrick Calvin Wamala}
\IEEEauthorblockA{\textit{Dept. of Electrical and Computer Engineering} \\
\textit{Makerere University}\\
rodrickwamala@gmail.com}
\and
\IEEEauthorblockN{Jonathan Serugunda}
\IEEEauthorblockA{\textit{Dept. of Electrical and Computer Engineering} \\
\textit{Makerere University}\\
seruguthan@gmail.com}
\and
\IEEEauthorblockN{* Andrew Katumba}
\IEEEauthorblockA{\textit{Dept. of Electrical and Computer Engineering} \\
\textit{Makerere University}\\
katraxuk@gmail.com}
		
}

\maketitle
\begin{abstract}
\footnote{Corresponding author. katraxuk@gmail.com}
Online information systems currently heavily rely on the username and password traditional method for protecting information and controlling access. With the advancement in biometric technology and popularity of fields like AI and Machine Learning, biometric security is becoming increasingly popular because of the usability advantage. This paper reports how machine learning based face recognition can be integrated into a web-based system as a method of authentication to reap the benefits of improved usability. This paper includes a comparison of combinations of detection and classification algorithms with FaceNet for face recognition. The results show that a combination of MTCNN for detection, Facenet for generating embeddings, and LinearSVC for classification out performs other combinations with a 95\% accuracy. The resulting classifier is integrated into the web-based system and used for authenticating users.\par

\end{abstract}

{\bf \textit{\small{Keywords--FaceNet, MTCNN, Face Recognition, Machine Learning, Biometric Authentication, LinearSVC}}} 

\section{Introduction}
Information and Communication Technology usage has witnessed rapid growth in the past decade
decade all around the world. A bigger percentage of the population has laptops, personal computers, and smart phones making it easy to access the internet and thus changing lives of millions of people \cite{b01}.
All web-based systems that have users and store personal information about the users require a mechanism to keep track of their users’ information. Most commonly every user of the system is assigned an instance in the database that represents them (their identity). To protect the user identity, an authentication and authorization mechanism is implemented to control access to certain information. The most common method in web-based systems is authentication using passwords. Users regularly provide a combination of their username and password  through a form to access a remote account \cite{b1}.\par
Passwords have well-known disadvantages in both usability and security. This leads to promotion of biometric-based authentication. The primary motivation of biometric authentication is usability: users are not required to remember the passwords, there is nothing for them to carry, biometric systems are generally easy to use, and scalable in terms of the burden exerted onto the users. Biometric technology can be used to control the risk of sharing, forgetting, losing, and embezzlement of passwords \cite{b2}.\par
\subsection{Biometric Authentication}
Biometric authentication is the automatic authentication that identifies a person by analyzing their physiological and/or behaviour features. In the past two decades, both physiological and behaviour characteristics have been used in biometric authentication systems. Physiological features are attributed to the person's body shape such as the face, palm prints, DNA, e.t.c. Behavioural characteristics are attributed to the person’s behavior: typing speed, voice, and articulation \cite{b3}.\par
\subsection{Artificial Intelligence}
Artificial Intelligence is impacting the world with its ability to achieve the never thought tasks, the accurate resource knowledge used can highly affect the execution. In recent years we all have witnessed an exceptional growth in the improvement of state of the art accuracy for multiple domains such as text classification, speech recognition, image processing and natural language processing \cite{AI1}.\par The challenges faced by hard-coded systems convey that artificial intelligence systems require an ability to extract and acquire knowledge from raw data supplied to them. The idea of systems learning from data, identifying patterns and making decisions with little human intervention is known as machine learning. Real world AI applications face a difficulty caused by the fact that the factors of variation influence every unit of data that can be observed. For example, individual pixels in a blue object image may be very close to those of a black object at night. The shape of the object depends on the angle of observation. Most of the applications require us to extract the most important factors and discard the ones we do not care about. Deep learning, a subset of machine learning implements this by introducing factors represented in terms of other and simpler factors. Deep learning allows the computer to build complex problems out of simpler problems utilizing a hierarchy of artificial neural networks to carry out machine learning processes. Artificial neural networks are inspired and built like the biological neural network of a human brain. It consists of neuron nodes connected together. Traditional programs develop analysis with data in a linear way, while the hierarchical style of deep learning systems gives machines the ability to process data with in nonlinear way. \cite{AI3}\cite{AI4}\cite{AI5}.

\subsection{Face Recognition}
In the present time, face recognition is a popular and  important technology integrated in many applications such as payment, door unlock, video monitoring systems, etc. A face recognition system is a technology that can identify and verify people from digital images and footage. The state-of-the-art face recognition approaches for web authentication include luxand \cite{b1} and smileid from electronicsid \cite{b2}. These are both provided as commercial products. \cite{b3-i}. To recognize faces (human faces), Images are taken using a digital camera. the first step is to perform face detection, this is performed as a preprocessing step to locate the face in an image. Faces are generally unique depending on the environment the images were taken: illumination, brightness, facial expressions, and the person's face orientation. The next step is the face recognition: a set of features are computed utilizing a special algorithm. The set of features generated is classified to represent each person’s face uniquely. The response from the classifier is the owner of the facial features \cite{b4}.\par
A number of face recognition studies are carried out but these disclose only the pipeline representation stage. This was started years ago by the mathematical background. In any case, from the academic perspective, researchers from the Carnegie Mellon and Oxford universities have shared both network structures and pre-trained models. From the commercial perspective, companies (Google and Facebook) share only the network structures. Lucky for us there is an encouraging open source community that has emerged who train these models from scratch and share them. \cite{b5}.\par
There are many face recognition algorithms in the academic field, part of the research results are relatively mature. The state of the art face recognition models include VGGFace \cite{b6} from the University of Oxford, DeepFace\cite{b7} from Facebook, OpenFace \cite{b8} from the Carnegie Mellon University, and FaceNet \cite{b9} from Google.\par
A study that compares the four state-of-the-art recognition models was conducted during the development of the lightFace \cite{b5} framework (A python framework that provides face-recognition support for each of these models). It showed that Google’s Facenet was the best model among the four.\par

\section{Implementation}
Our Approach on face recognition consists of three stages, that is given an image, we perform face localization, face embeddings generation using Google’s FaceNet pretrained model, and classification using the linearSVC classifier.\par

\subsection{Data Collection}
In a case study on 100 different people, a total of 1424 images from the volunteers. Captured using a 12MP dual-lens camera of the tecno camon 12 air smartphone, the images appeared in different orientations and lightings. The number 1424 is that which was obtained after performing data cleaning.\par

\subsection{Hardware Setup}
The training was done on a personal computer. A HP ProBook 450 G3, with 8192MB RAM,
Intel(R) Core(TM)i5-6200U CPU @ 2.30GHz(4 CPUs) 2.4GHz processor, and AMD Radeon
(TM) R7 M340 graphics card with 2GB dedicated VRAM.

\subsection{Face Detection}
For the case of face detection, two of the most commonly used ways of face detection were compared;
\subsubsection{The Viola-Jones Haarcascade classifier}
It is based on the Viola-Jones Object Detection framework. This frame work has a quiet high (true positive rate) detection rate and a very low false positive rate making the algorithm a robust one that also processes the images quickly. The main objective is face detection not recognition: it distinguishes faces from non-faces which is the first (preprocessing) step recognition. The framework follows four main steps: Haar Feature Selection, Creating an Integral Image, Adaboost Training and the Cascading Classifiers.\cite{b10}\par

\subsubsection{The Multi-Task Cascaded Convolutional Neural Network} that is based on Neural Network-Based Face Detection. A Neural Network inspired by human brain composed of simple artificial neurons also known a perceptrons are connected to each other in multiple layer. The MTCNN is comprised of three layers.\par

\begin{figure}[H]
  \centering
  \includegraphics[scale=0.45]{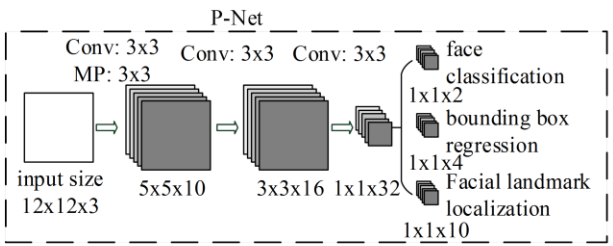}
  \caption{The Proposal Network Layer \cite{b12}}
  \label{fig:pnet}
\end{figure}

Layer 1: \emph{The Proposal Network (P-Net): } This is a fully convolutional network which is used to obtain the candidate windows and their bounding box regression vectors as shown in figure~\ref{fig:pnet} \cite{b12}

Layer 2: \emph{The Refine Network (R-Net): } The output of the P-Net is input into the R-Net which is a Convolution Neural Network and hence denser than the P-Net layer. The R-Net further reduces the number of candidates (which further rejects a large number of false candidates)\cite{b12}, performs calibration with bounding box regression and employs Non-Maximum Suppression (NMS) to merge overlapping candidates.

\begin{figure}[H]
  \centering
  \includegraphics[scale=0.38]{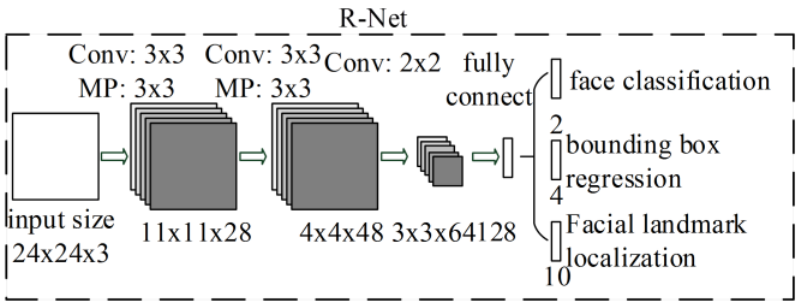}
  \caption{The Refine Network Layer \cite{b12}}
  \label{fig:rnet}
\end{figure}

Layer 3: \emph{The Output Network (O-Net): } Similar to the R-Net the O-Net aims to describe the face in detail outputting five facial landmarks' positions for the main face features like the eyes, nose and mouth.


Face detection and localization was performed on the faces using the MTCNN algorithm. The localized face is resized into a 160×160 RGB image. A face dataset is created.\par

\subsection{Face Recognition}
A FaceNet model pre-trained on a VGGFace2 dataset consisting of 3.3M faces and 9000 classes in an inception ResNet v1 architecture \cite{b13} was used. The pretrained model expects 160×160 RGB images. The model was converted into a keras compatible version \cite{b14}. The converted model was used to generate embeddings for all images in the faces dataset. For each face image in the faces dataset, an array of 512 different weights uniquely representing the face was generated. A new dataset of face embeddings was prepared and compiled for classification purposes.
The generated embeddings are used to create a new dataset consisting of face embeddings. Table \ref{tab:model} below shows information about the pretrained Facenet model.

\begin{table}[H]

\centering 

\caption{Pre-trained FaceNet Model}
\scalebox{0.82}{
   \begin{tabular}{| l| l | l | l |}
	\hline
	 \hline
	  Model name & LFW accuracy & Training dataset & Architecture\\ \hline
	  20180402-114759 & 0.9965 &VGGFace2 & Inception ResNet v1 \\ \hline
    \end{tabular}
    }
\label{tab:model}
\end{table}

\subsection{Classification}
With the face embeddings dataset, the next step was classification of the face embeddings. Classification involved comparison of three classification algorithms: the LinearSVC, OneVsRest Classifier, and the Random Forest Classifier.\par
Before training the classification models, the embeddings dataset was split into the test and train datasets, $30\%$ for testing, and $70\%$ for training.  For each classification model trained, a 10-fold stratified
cross-validation method was used for tuning the classifier to obtain improved performance. The train set was made up of 952 images, the test set 472 images and the validation fold 95 images.\par

After training each of the classification algorithms with the train data (70\% of 1424 face embeddings dataset), each trained classification model was tested by tasking it to predict the test
data.
Each of the classifiers was validated using the stratified (10-fold) cross-validation method. The
individual accuracies for different folds were determined to reflect the performance of the model
at different sections for the entirety of the training dataset. For each fold, the validation dataset
was 10\% of the training dataset.

\subsubsection{LinearSVC}
The classification model here was trained with a regular parameter, $randomstate = 0$, and tuned with a hyperparameter $C=1$. It took approximately $14 seconds$ to train this model, the average validation accuracy was 0.93.
\subsubsection{OneVsRest}
This classifier used the SVC with a regular parameter $probability = True$, as the binary classifier for performing multiclass classification. The model was tuned with a hyperparameter, $n\_jobs = 100$. It took approximately $97 seconds$ to train this model on the train set, and the average cross-validation accuracy was 91\%. 
\subsubsection{Random Forest}
The classification model was created with a regular parameter, $randomstate = 0$, and tuned with a hyperparameter $n\_estimators = 1000$. The training for this model took approximately $120 seconds$. The average validation accuracy was $91\%$.\par

Each of the classification algorithms was evaluated with metrics: precision, recall, accuracy, and the confusion matrix to visualize wrong classifications.

\subsection{Bias Test}
Two datasets of the same size with images from 33 different people each were generated. One dataset consisted of entirely white people's images obtained from the Georgia Institute of Technology \cite{b11}m the other consisted of entirely black people's images. It was a fraction of the data collected from the 100 different volunteers. Face embeddings from the respective datasets were classified and compared, the performance metrics to determine if there is bias in the pre-trained model.\par
\begin{figure}[H]
  \centering
  \includegraphics[scale=0.27]{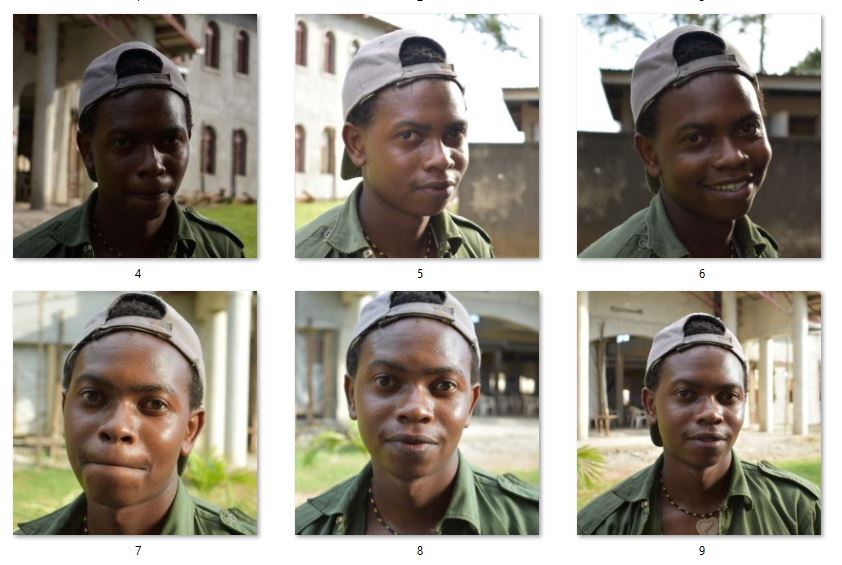}
  \caption{Black people dataset sample}
  \label{fig:mtcnndec}
\end{figure}
\begin{figure}[H]
  \centering
  \includegraphics[scale=0.35]{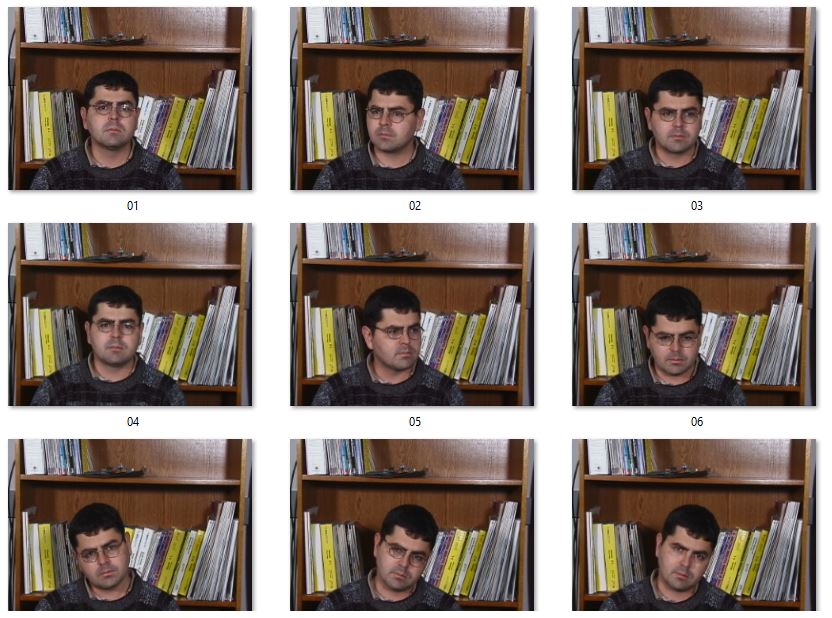}
  \caption{White people dataset sample}
  \label{fig:mtcnndec}
\end{figure}

\section{Face Authentication}
A login view that captures the person’s face at the frontend was developed. This took the image and converted it into a data URI. At the face recognition login view, there is an event listener on the login button that makes a call to the face recognition API with a request composed of the image data URI. When the recognition API receives the request, it extracts the URI converts it into an image and fowards it to the classification model to perform recognition. The model responds with a code corresponding to the person found in the image. The API then sends the predicted code back to the frontend. The frontend sends this code to the system backend RESTful API. In the web-based system, as a user registers for face recognition, the system encrypts, and stores their special code in the database. When the frontend/client receives the predicted code from the face recognition API, It makes a call to the system backend RESTful API with a request composed of the unique code. The system then extracts the code and user-email, decrypts the code it has in the database under the corresponding user-email, and checks whether the two match. If they match, the user is authenticated and authorized into their account, otherwise, authentication fails.\par
The stack used for system development was VueJS at the frontend, expressJS for the backend API, MongoDB for data storage, and python's flask for the recognition API.\par
\begin{figure}[H]
  \centering
  \includegraphics[scale=0.18]{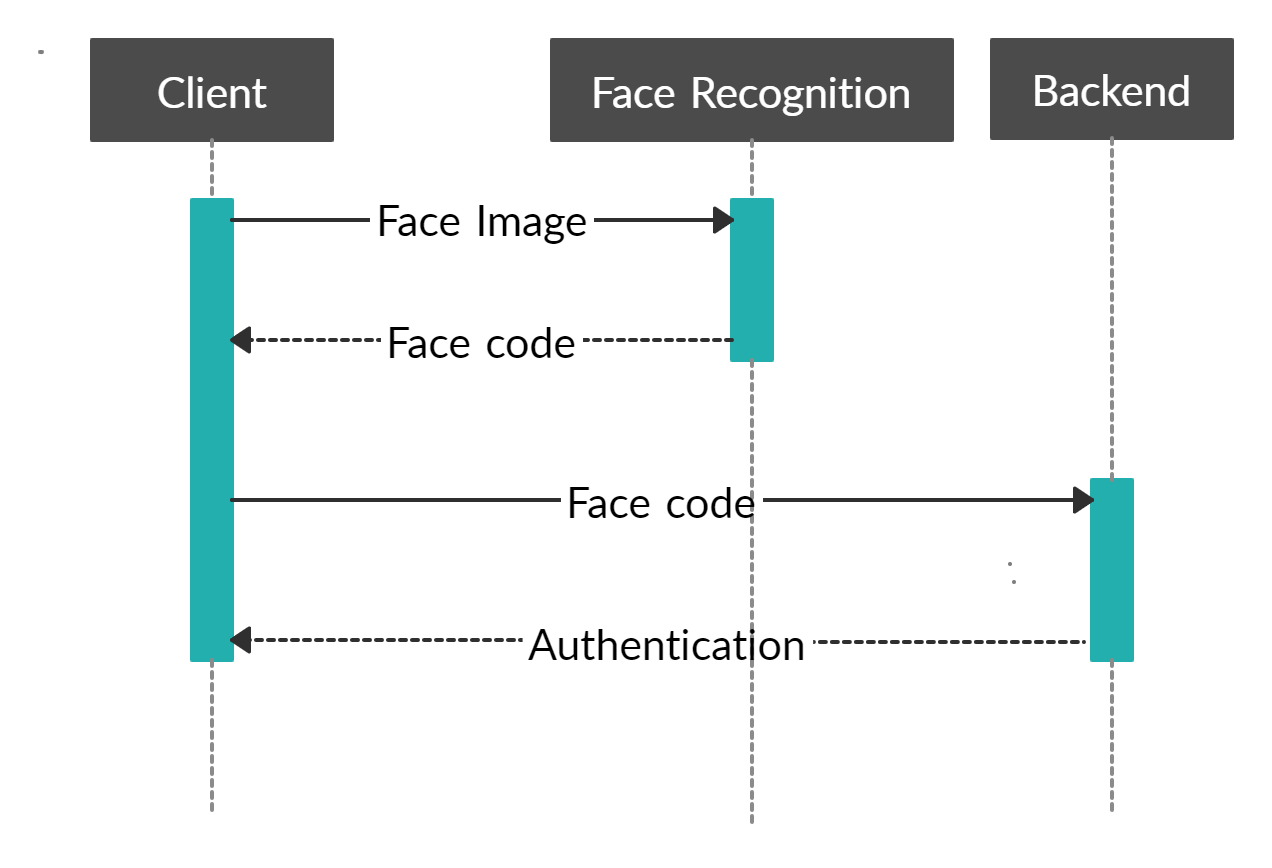}
  \caption{Face authentication sequence diagram}
  \label{fig:mtcnndec}
\end{figure}

\section{Results}
\subsection{Face Localization}
\begin{figure}[H]
  \centering
  \includegraphics[scale=0.6]{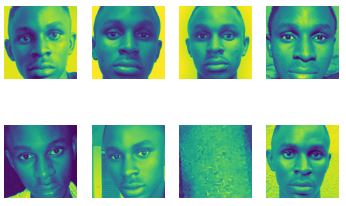}
  \caption{Localization using the Viola-Jones Haarcascade algorithm}
  \label{fig:lochaar}
\end{figure}

Face localization using the Viola-Jones Haarcascade algorithm produced results in figure \ref{fig:lochaar}. It can be observed that the algorithm eliminated some of the vital facial features like the chin, which in turn affected the accuracy of recognition with a face dataset generated using this algorithm. The algorithm failed to localize faces despite existence in the image on several occasions, for example in figure \ref{fig:lochaar}, the localization in the second row third image is faceless.\par
\begin{figure}[H]
  \centering
  \includegraphics[scale=0.6]{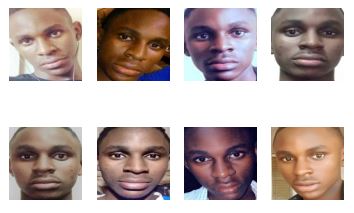}
  \caption{Localization using the MTCNN algorithm}
  \label{fig:locmtcnn}
\end{figure}

The MTCNN algorithm, on the other hand, could perfectly detect a face whenever one existed in the image. The results of the MTCNN detection on a similar group of images can be seen in figure \ref{fig:locmtcnn}. By observation, the MTCNN out performs the Viola-Jones algorithm in localizing faces.\par
When the two algorithms were used to generate identical datasets, which were further used to train the linearSVC model, the performance metrics in table \ref{tab:facedectable} below were obtained. A $74\%$ versus $98\%$ accuracy on a similar dataset reflects how good a choice the MTCNN is compared to the Viola-Jones haarcascade.
\begin{table}[H]
\centering 
\caption{comparison of the face detection algorithms metrics} 
   \begin{tabular}{| l | l | l |}
	  \hline
	  \hline
	  Algorithms & Accuracy & Precision\\ \hline
   	  Haarcascade &  0.74  & 0.74 \\ \hline
	  MTCNN & 0.98 & 0.98 \\ \hline
    \end{tabular}
\label{tab:facedectable}
\end{table}
Table \ref{tab:facedectable} shows the results obtained from the comparison of the two face detection algorithms we considered in our research basing on the accuracy and precision as metrics of performance.

\subsection{Classification}
The performance metrics for classification were as indicated in the table \ref{tab:classificationtable}.

\begin{table}[H]
\caption{Classification algorithms metrics} 
\centering 
   \begin{tabular}{| l | l | l | l | l |}
	 \hline
	\hline
	 Classifier & Precision & Accuracy & Recall & F1 Score\\ \hline
	  LinearSVC & 0.95 & 0.95 & 0.95&0.95 \\ \hline
	  OneVsRest & 0.94 & 0.94&0.94 & 0.93 \\ \hline
	  Random Forest & 0.91 & 0.91 & 0.91&0.91\\ \hline
    \end{tabular}
\label{tab:classificationtable}
\end{table}

\subsection{Model Selection}
\begin{figure}[H]
  \centering
  \includegraphics[scale=0.38]{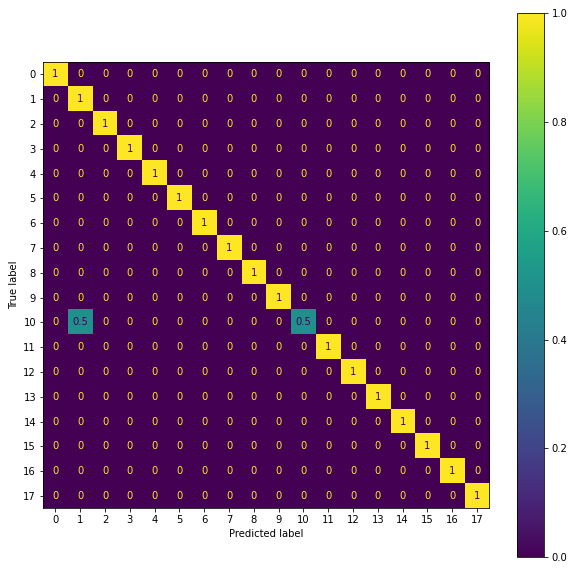}
  \caption{LinearSVC confusion matrix for 20 of the 100 classes}
  \label{fig:cfm}
\end{figure}

From the results discussed above, it was determined that the LinearSVC provided a better
performance by 1\% and 4\% compared to the OneVsRest and Random Forest classifiers respectively. Thus the LinearSVC model was exported and utilized in the development of the face
recognition API.
Figure \ref{fig:cfm} below is a normalized confusion matrix for 20 of the 100 classes (people) classified using
the LinearSVC classifier. From this figure, it can be determined that the class labeled 10 was
wrongly classified as the class labeled 1 for half of its data items in the test set.
\subsection{Bias Test}
Table \ref{tab:dcomp} shows the results obtained from the comparison of the performance of the linearSVC when trained on a black faces dataset compared to a white faces dataset.

\begin{table}[H]

\centering 
\caption{Datasets Comparison} 
   \begin{tabular}{| l |  l | l |  l | l |}
	 \hline
	\hline
	  Dataset & Precision & Accuracy& Recall & F1 Score\\ \hline
	  Black people & 0.94 & 0.94 & 0.93 & 0.92 \\ \hline
	  White people & 1.00 & 1.00 & 1.00 & 1.00\\ \hline
    \end{tabular}
\label{tab:dcomp}
\end{table}
The metrics reflect the fact that the model produced a better performance with a white faces dataset compared to a similar size blackfaces dataset. Thus the pretrained face recognition model used to generate face embeddings that uniquely represent the face had bias toward black faces.

\subsection{System}
Figure \ref{fig:facereclogin} below displays the face-recognition login interface that the user sees when they choose the login with face recognition option from the system interface.

\begin{figure}[H]
  \centering
  \includegraphics[scale=0.25]{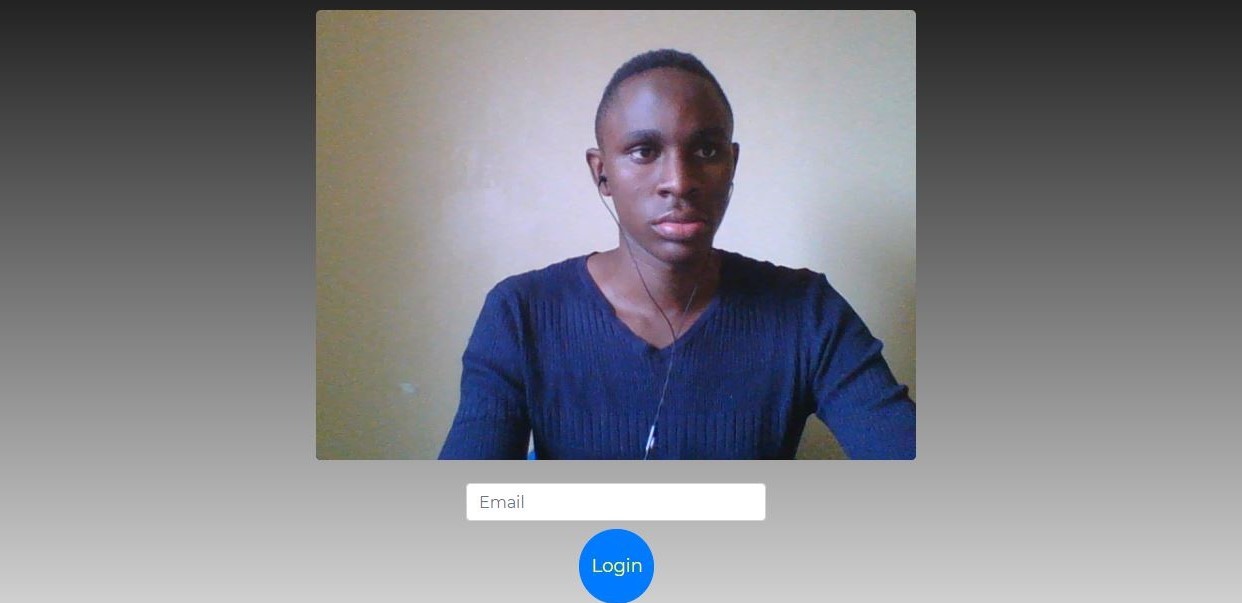}
  \caption{Face recognition login interface}
  \label{fig:facereclogin}
\end{figure}

\section{Conclusion}
The system developed uses the MTCNN algorithm to detect the face, generates embeddings using a FaceNet pretrained model and a linearSVC classifier to recognize images taken through the system and authenticates users accordingly. The combination of these algorithms resulted in a 95\% recognition accuracy.
The limitations for this work include the fact that it was determined that the pre-trained FaceNet model used had bias towards the black people dataset and the issue of liveness at the login interface, which made it possible for one to login into the system using an image since there is no mechanism for liveness detection. With these limitations, we propose as our future work to retrain the FaceNet model with a black faces dataset to eliminate bias, implement classfication with neural networks, integrate liveness detection at the frontend of the system face recognition login, automate the process of classification for digital onboarding of new users.\par


\end{document}